\documentclass[runningheads]{llncs}
\usepackage[utf8]{inputenc}
\usepackage{graphicx}
\usepackage[table,xcdraw]{xcolor}
\usepackage{url}
\usepackage{float}
\usepackage{amsmath}
\usepackage{multirow}
\usepackage[caption=false]{subfig}
\usepackage{comment}
\usepackage{multicol}
\usepackage{cite}

\newcolumntype{L}[1]{>{\raggedright\let\newline\\\arraybackslash\hspace{0pt}}m{#1}}
\newcolumntype{C}[1]{>{\centering\let\newline\\\arraybackslash\hspace{0pt}}m{#1}}
\newcolumntype{R}[1]{>{\raggedleft\let\newline\\\arraybackslash\hspace{0pt}}m{#1}}
\begin{document}
\title{An Embedded Monocular Vision Approach for Ground-Aware Objects Detection and Position Estimation\thanks{Supported by Centro de Informática (CIn - UFPE), Fundação de Amparo a Ciência e Tecnologia do Estado de Pernambuco (FACEPE), and RobôCIn Robotics Team.}}
%
\titlerunning{Monocular Vision Object Position Estimation}
%
\author{
João G. Melo\and 
Edna Barros}
\authorrunning{J. Melo et al.}
%
\institute{Centro de Informática, Universidade Federal de Pernambuco. \\
Av. Prof. Moraes Rego, 1235 - Cidade Universitária, Recife - Pernambuco, Brazil.
\email{\{jgocm, ensb\}@cin.ufpe.br}}

\authorrunning{J. Melo et al.}
%
\maketitle              

\begin{abstract}

In the RoboCup Small Size League (SSL), teams are encouraged to propose solutions for executing basic soccer tasks inside the SSL field using only embedded sensing information. Thus, this work proposes an embedded monocular vision approach for detecting objects and estimating relative positions inside the soccer field. Prior knowledge from the environment is exploited by assuming objects lay on the ground, and the onboard camera has its position fixed on the robot. We implemented the proposed method on an NVIDIA Jetson Nano and employed SSD MobileNet v2 for 2D Object Detection with TensorRT optimization, detecting balls, robots, and goals with distances up to 3.5 meters. Ball localization evaluation shows that the proposed solution overcomes the currently used SSL vision system for positions closer than 1 meter to the onboard camera with a Root Mean Square Error of 14.37 millimeters. In addition, the proposed method achieves real-time performance with an average processing speed of 30 frames per second.

\end{abstract}

\section{Introduction} \label{intro}
At the RoboCup Small Size League (SSL) robot soccer competition, games occur between two teams of omnidirectional mobile robots with eight players for division A and six for division B. Frames from cameras placed above the field are processed by a dedicated computer, which runs SSL Vision: a standard vision system for detecting and tracking elements such as robots, goals, balls and field lines \cite{ssl-vision}. Off-field computers, one for each team, receive the position information and referee commands and perform most of the computation and exchange of information with robots using Radio Frequency (RF) communication with minimal bandwidth.

In recent RoboCup editions, the League has proposed a new technical competition in which teams are only allowed to use embedded sensing information for executing basic soccer tasks inside the field \cite{vision-blackout}. The Vision Blackout challenge encourages teams to propose autonomous navigation solutions for the SSL environment. In 2021 the teams were assigned three tasks for the competition: grabbing a stationary ball somewhere on the field (1), scoring with the ball on an empty goal (2), and scoring with the ball on a statically defended goal (3). Therefore, detecting and locating balls, robots, and goals from embedded devices is needed.
 
SSL robots are constrained to a 180mm diameter limit and achieve up to $3.7$ m/s velocities, requiring low-power, small-size, high-throughput solutions. For example, SSL ball detection using scan lines and color segmentation has been previously proposed \cite{tigers-2019}. However, even though this approach has achieved accurate results in recent competitions, it cannot detect other SSL objects. Moreover, it lacks robustness concerning local illumination or field changes.

With the advances in Deep Neural Networks (DNN) architectures and parallel processing technologies, the use of Convolutional Neural Networks (CNN) for
object detection has grown considerably in embedded applications \cite{edge-devices-benchmark}. This approach presents significant advantages compared to traditional computer vision techniques, especially in robustness to environmental conditions and adaptability to new object classes. For that, an open-source SSL dataset\footnote{https://github.com/bebetocf/ssl-dataset} containing 2D bounding boxes for detecting balls, goals, and robots on images is available \cite{dataset-ssl}.

Computing position and orientation (pose) from detected objects is also essential for autonomous navigation. Considering the robot's resource-constraints, solutions employing a single monocular camera for object detection and position estimation are preferred. Exploiting prior knowledge from the environment, Inverse Perspective Transformation can be applied for computing three dimensional positions from camera frames \cite{ground-localization}.

This research proposes a monocular vision solution for detecting and estimating the relative positions of SSL objects, using bounding boxes' 2D coordinates from a CNN-based Object Detection model and previously calibrated camera parameters for back-projecting objects positions on the soccer field. The proposed solution is tested on an NVIDIA Jetson Nano Developer Kit \cite{jetson-nano} employing a Logitech C922 camera, both mounted on the top of a SSL robot, and we evaluate accuracy results for ball localization, achieving a 14.37 millimeters Root Mean Square Error (RMSE), overcoming position accuracy from the standard SSL Vision system. The implemented system runs at 30 frames per second, with an average 10.8 Watts power consumption, and respects all League's restrictions, showing it can be applied for the desired challenge, and the main contributions of this work are:

\begin{itemize}
    \item A complete architecture for detecting and locating SSL objects using CNN-based object detection.
    \item Presenting procedures for calibrating onboard camera intrinsic and extrinsic parameters.
    \item A detailed pipeline and configurations for training SSL Object Detection and deploying to NVIDIA Jetson Nano.
\end{itemize}

\section{Related Work}\label{ch2}

The SSL Vision Blackout Challenge was introduced in the League in 2019, with Tigers Mannheim achieving the best results \cite{tigers-2019}. Their work aims to detect the ball and estimate its relative position to the robot and their research reports that CNNs may achieve robust results \cite{OKeeffe2017ABD}, but the team has discarded its use due to the lack of embedded hardware accelerators by the time; blob detection methods are too slow and highly sensitive to lighting changes; an edge detection method took too much processing time as well and identified moving balls as ellipsoids. Therefore, Tigers comes up with a novel approach: an algorithm that searches for ball candidates by applying a sharp edge detection kernel along logarithmically distributed horizontal lines. A vertical scan is performed when candidates are found, and the ball's radius, confidence, and position are estimated. The team also shares a deployment infrastructure for the on-bot vision software \cite{tigers-2021} and a full architecture for autonomous robot-ball interaction \cite{tigers-2020}, on which ball distances to the robot are estimated from the object size on the image.

At other RoboCup soccer leagues, Deep Learning methods are proposed for
detecting elements such as balls, robots, goalposts, and field lines. At the
Standard Platform League, for instance, CNNs are used for detecting field
boundaries \cite{field-boundary-cnn}, robots and balls, while also estimating
ball's positions \cite{DeepRobotDetection}. Since
detecting goals and field lines are essential for self-localization on the
soccer field, another approach defines four object classes for the network to
detect: balls, line crossings, robots, and goalposts \cite{ROBO}. Exploiting
prior knowledge about the ball characteristics, another research presents an
algorithm for searching ball regions proposal to accelerate inference time by
applying the object detection CNN to a smaller region of the image
\cite{cnn-ball-detection}.

An open-source dataset containing 2D bounding boxes labels for SSL robots, balls
and goals was introduced in 2021's RoboCup edition \cite{dataset-ssl}. The
research also benchmarks Deep Learning state-of-the-art object detection models
evaluating their accuracy and inference speed on a Google Coral TPU accelerator,
reporting SSD MobileNets V1 and V2 to achieve the best overall performances.

The main bottleneck for using CNNs on SSL was the lack of hardware-accelerated
embedded platforms for running DNNs in real-time \cite{tigers-2020}. With Deep
Learning empowering several IoT applications, the concept of Edge Computing
rapidly gains huge attention, urging for low power devices capable of running
complex DNNs \cite{edge-computing}. Edge Devices comparisons show that NVIDIA
Jetson Nano and Google Coral Developer Board achieve the best overall results
when running object detection models, concerning the accuracy, inference time
and power consumption \cite{edge-devices-benchmark}.

As minor performance improvements are extremely relevant in real-time
applications, numerous CNN architectures are proposed in the object detection
domain, is mostly divided into two categories: two-stage and one-stage, such as single
shot detectors (SSD). For mobile applications, single-shot ones gain most of the
attention and previous comparisons between state-of-the-art models report SSD
MobileNet V2\cite{ssd, mobilenetv2} to achieve great speed-accuracy
trade-offs, especially under hardware constrained scenarios
\cite{object-detection-constrained-hardware, dataset-ssl, nvidia-benchmark}.

As 2D object detection models are only capable of computing two-dimensional
bounding box positions on images, we also search for three-dimensional position estimation solutions. A DNN combining human semantic segmentation and depth prediction achieves remarkable accuracy with high-speed inferences on Jetson Nano \cite{human-depth-estimation}. However, the network must be trained with a depth-labeled dataset, which is not available for SSL objects.  At the RoboCup@Home League, OpenPose CNN architecture \cite{open-pose} is used for detecting key points, and object poses are calculated from 2D-3D correspondences between detected and ground-truth key points using PnP-RANSAC \cite{6d-pose-estimation}. However, the method presents limitations when dealing with symmetric objects. Also, works on vision-based self-localization show that relative positions from points on images can be retrieved by Inverse Perspective Transformation if one of its world coordinates is known \cite{ground-localization}.

In this work, we solve the Perspective-n-Point (PnP) problem for 2D-3D correspondences between a set of hand-marked points on the field for estimating a camera's relative pose to the field coordinates. Also, we employ SSD MobileNet v2 for regressing object's 2D bounding boxes and a linear regression model for calculating their bottom-centers projection on camera frames. Extrinsic and intrinsic parameters are used for solving the Inverse Perspective Transformation problem for points on the ground and object's relative positions are estimated.

\section{Proposed Approach}\label{ch3}

\begin{figure}[ht]
    \begin{center}
        \includegraphics[width=0.95\textwidth]{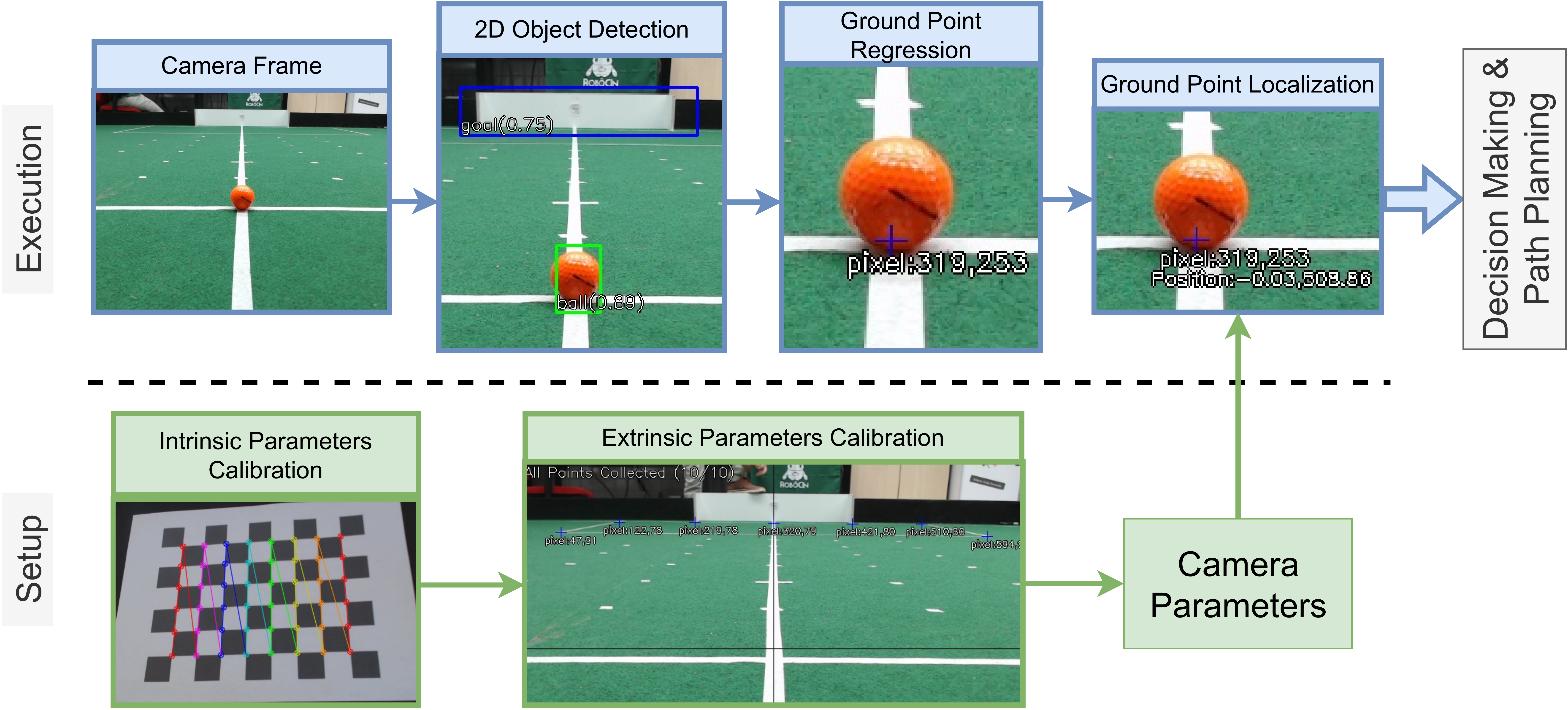}
    \end{center}
    \caption{Designed architecture for detecting and locating SSL objects with a single monocular camera.}
    \label{fig:pipeline}
\end{figure}

During soccer matches and especially for the Vision Blackout challenge, SSL objects mostly lay on the soccer field, and we exploit this prior knowledge for proposing a monocular vision solution for detecting and estimating their relative positions to the robot. For that, the camera is fixed to the robot and its intrinsic and extrinsic parameters are obtained using calibration and pose computation techniques from the Open Computer Vision Library (OpenCV) \cite{opencv-library}. A state-of-the-art CNN-based object detection model, SSD MobileNet v2 \cite{ssd, mobilenetv2}, is used for detecting objects on camera frames. After labeling, linear regression is applied to the bounding box’s coordinates, assigning a point on the field that corresponds to the object’s bottom center, which has its relative position to the camera, and, therefore, to the robot estimated using pre-calibrated camera parameters. Position information can be delivered to decision-making and path planning algorithms to execute autonomous navigation. Figure \ref{fig:pipeline} illustrates a scheme for the proposed method and all software is open-source\footnote{https://github.com/jgocm/ssl-detector}.

In the following subsections, in-depth explanations for each of the steps from the proposed pipeline for object localization are presented. Firstly, we depict the camera pinhole model and calibration procedures. Then, in sequence, since the proposed approach for estimating objects' positions is based on ground points localization, a method for computing field points' relative positions is presented. Then, CNN model conversion and training details are given, followed by an explanation of the procedure adopted for fitting a linear regression model for estimating the object's ground pixel.

\subsection{Camera Calibration}
\label{sec:camera-calibration}
Camera parameters are divided into intrinsic and extrinsic, and the process for projecting three-dimensional points to the image plane can be described in three steps: converting 3D world position to the camera coordinates system using the extrinsic parameters (1); projecting points to the image plane using intrinsic parameters (2); and re-scaling pixels using a scale parameter (3). The pinhole camera model describes the mathematical representation for this process \cite{tigers-2020}, which is given in detail by Equation \ref{eq:pinhole} and simplified on Equation \ref{eq:pinhole-simple}.

\begin{equation}
    \label{eq:pinhole}
    s\:
    \begin{bmatrix} u \\ v \\ 1 \end{bmatrix}
    =
    \begin{bmatrix}
        \alpha_{x} & \gamma     & u_{0} \\
        0          & \alpha_{y} & v_{0} \\
        0          & 0          & 1
    \end{bmatrix}
    \begin{bmatrix}
        r_{11} & r_{12} & r_{13} & t_{1} \\
        r_{21} & r_{22} & r_{23} & t_{2} \\
        r_{31} & r_{32} & r_{33} & t_{3}
    \end{bmatrix}
    \begin{bmatrix} x_{w} \\ y_{w} \\ z_{w} \\ 1 \end{bmatrix}
\end{equation}

\begin{equation}
    \label{eq:pinhole-simple}
    s\:p_{c}=K\;[R\:|\:t]\;p_{w}
\end{equation}

In Equation \ref{eq:pinhole-simple}: $p_{w}$ represents a 3D point in the world coordinates system; $[R\:|\:t]$ describes the camera axis rotation and
translation concerning the world coordinates, which are called extrinsic
parameters; $K$ is the intrinsic parameters matrix, consisting of $\alpha_{x}$ and $\alpha_{y}$ scale factors, $u_{0}$ and $v_{0}$ coordinates for the principal point and the $\gamma$ skew factor; $p_{c}$ is the pixel position on screen; and $s$ is a depth scale factor. Thus, for back-projecting image pixels to three-dimensional world coordinates, camera parameters must be calculated beforehand.

\begin{figure}[H]
    \begin{center}
        \includegraphics[width=0.4\textwidth]{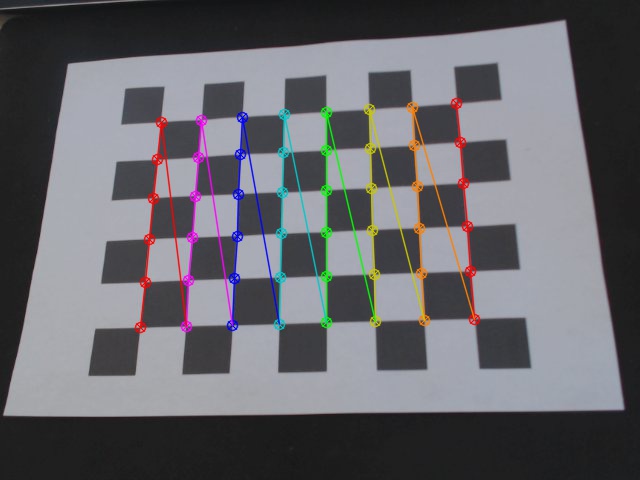}
    \end{center}
    \caption{Corner detection and chessboard pattern recognition result. The
        method is applied to multiple images and, by the end, the 2D points and its 3D correspondences are used for estimating the camera's intrinsic parameters and lens distortions \cite{camera-calibration}.}
    \label{fig:chessboard-calibration}
\end{figure}

The chosen procedure for estimating intrinsic camera parameters requires
multiple images of a planar pattern from different viewpoints, making 2D-3D
correspondences between pattern points from different images
\cite{camera-calibration}. For instance, a chessboard pattern was used and
OpenCV implementations for camera calibration and chessboard corner detection
for pattern recognition were applied, as illustrated in Figure
\ref{fig:chessboard-calibration}.

\begin{figure}[ht]
    \begin{center}
        \includegraphics[width=0.7\textwidth]{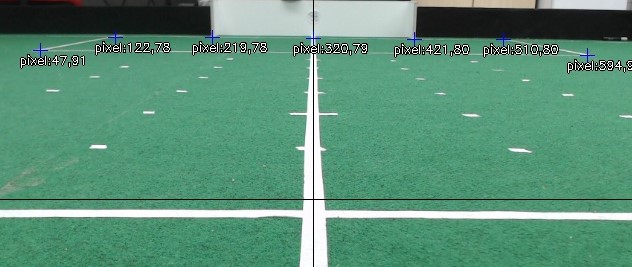}
    \end{center}
    \caption{With the camera mounted onto the robot, pixels corresponding to field points with known positions are hand-marked, for instance, penalty area corners, field corners, goal corners, or the goal center can be used. A PnP problem for the 2D points and their three-dimensional correspondences is solved by minimizing re-projection error. The employed algorithm is an implementation from OpenCV library and computes the camera`s rotation and translation vectors with respect to field coordinates.}
    \label{fig:field-calibration}
\end{figure}

For extrinsic parameters estimation, 2D-3D correspondences between SSL field
points can be used to find the camera rotation and translation regarding field axis by solving a Perspective-n-Point (PnP) problem, which consists of finding a camera's pose based on a set of pixels and its corresponding three-dimensional world coordinates. Figure \ref{fig:field-calibration} describes the calibration procedure.

\subsection{Ground Point Localization}
By using the calibrated extrinsic and intrinsic parameters, given a pixel on the screen, the camera model can be rewritten as Equation \ref{eq:pinhole-inverse}. From that, if rotation matrix $R$, translation vector $t$, and intrinsic parameters matrix $K$ are computed, any position $p_{w}$ can be retrieved if one of its coordinates is known. For example, points on the ground, which in case correspond to $z_{w} = 0$, can have their $x_{w}$ and $y_{w}$ coordinates estimated.

\begin{equation}
    \label{eq:pinhole-inverse}
    \begin{bmatrix} x_{w} \\ y_{w} \\ z_{w} \end{bmatrix}
    =
    s\:R^{-1}K^{-1}\begin{bmatrix} u \\ v \\ 1 \end{bmatrix}-R^{-1}t
\end{equation}

\subsection{Object Detection Model}
Object detection tasks require high computing power, being the most time-consuming step of our proposed architecture. Thus, the SSD MobileNet v2 model was selected for its better trade-off between accuracy and inference time, among other state-of-the-art
models \cite{nvidia-benchmark}.

With pre-trained weights on the COCO dataset, a model from the Tensorflow
framework was retrained for $57,565$ steps on a Google Colab
Notebook, running with a Tesla K80 GPU. The proposed dataset for SSL Object
Detection contains 931 images with up to 4182 instances of balls, robots, and goals labeled in Pascal VOC and YOLO formats \cite{dataset-ssl}. For the
training, images were randomly partitioned into train and test sets with an
$80/20$ proportion, and batch size was configured to $24$.

\begin{figure}[H]
    \begin{center}
        \includegraphics[width=0.7\textwidth]{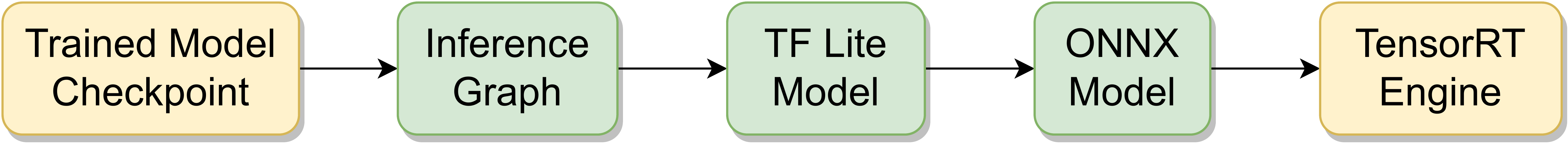}
    \end{center}
    \caption{The inference graph was exported from the Tensorflow retrained model checkpoint. TF Lite model was extracted from the graph and converted to ONNX format. NonMaxSupression (NMS) post-processing operation was replaced due to incompatibility issues, and the TensorRT Inference Engine was successfully generated from the ONNX file.}
    \label{fig:model-conversion}
\end{figure}

Using appropriate format conversion is essential for enabling high-performance hardware acceleration, especially on embedded platforms. For Jetson Nano, NVIDIA's TensorRT deep learning framework, which is built on CUDA parallel programming model, delivers low latency and high throughput for inference applications while also supporting models generated from Tensorflow, Pytorch, ONNX and other frameworks. Figure \ref{fig:model-conversion} gives an in-depth explanation for the model conversion procedure.

\subsection{Ground Point Linear Regression}

\begin{figure}[ht]
    \begin{center}
        \includegraphics[width=0.6\textwidth]{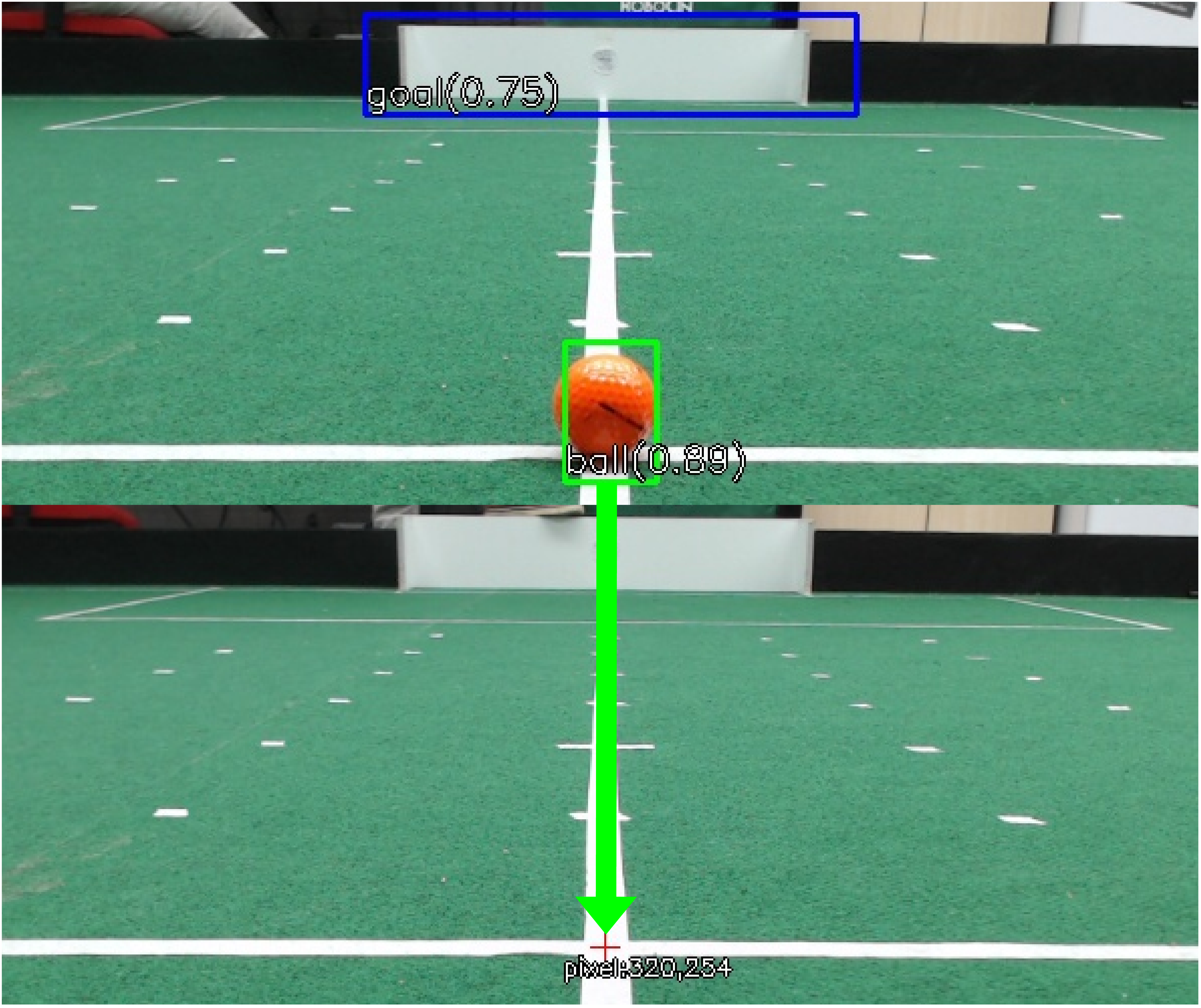}
    \end{center}
    \caption{With landmarks positioned with a 250mm grid on the field, the ball was placed on each of the markers and its bounding boxes were regressed from the object detection model. Then, removing the ball, the pixels that correspond to the center of each landmark on the screen were annotated. Thirty markers were used in the procedure.}
    \label{fig:linear-regression-fitting}
\end{figure}

In order to compute three-dimensional positions with the proposed method, we fit
linear regression weights for predicting a pixel that corresponds to the point
where the object touches the ground, that is $z_{w} = 0$. The inputs for the
model are bounding box coordinates generated from the 2D object detection
inference. Figure \ref{fig:linear-regression-fitting} illustrates the procedure
for finding the regression weights for the ball.

\section{Evaluation} \label{ch4}

The SSL uses standard golf orange balls, with an average $42.7$ millimeters diameter, for the soccer matches, and its localization was tested to evaluate the proposed method. Landmarks were positioned on the SSL field with a $250$ millimeters grid and used as the ground truth during the experiments. We placed the robot at a fixed position, setting camera XY coordinates to $0$ and $-500$. After calibrating camera intrinsic and extrinsic parameters, we estimated object relative positions on a set of images from the ball placed in 30 marked coordinates and compared the results to the SSL Vision system. 

For qualitative comparison and behavior analysis, ball XY coordinates were
plotted on Figure \ref{fig:scatter-plot}. As for quantitative measurements, Root
Mean Square Error (RMSE) was employed for the set of points taking landmark
positions as the ground truth. Angles between the objects and the robot are essential for navigating in the SSL environment and were also computed for evaluation. Coordinates measurements are in millimeters, while angles are in degrees.

\subsection{Camera Calibration Results}
Intrinsic parameters were estimated from 20 chessboard pictures taken in 640x480
resolution with a Logitech C922 camera, as presented in Section
\ref{sec:camera-calibration}. From the previously presented approach for
extrinsic parameters calibration, 5 points were hand-marked on the screen: the
bottom left and right goal corners, the lower left and right penalty area
corners, and the goal bottom center. Estimating rotation and translation vectors
from the PnP solution, camera to field axis relative pose can be computed. Calibration results are exhibited on Equation \ref{eq:calibration-results}.

\begin{equation}
    \label{eq:calibration-results}
    K
    =
    \begin{bmatrix}
        642.41 & 0      & 322.80 \\
        0      & 642.54 & 239.76 \\
        0      & 0      & 1
    \end{bmatrix},
    \;\;\;
    \begin{bmatrix}
        X \\
        Y \\
        Z
    \end{bmatrix}
    =
    \begin{bmatrix}
        -5.38   \\
        -509.79 \\
        171.40
    \end{bmatrix},
    \;\;\;
    \begin{bmatrix}
        \omega \\
        \phi   \\
        \kappa
    \end{bmatrix}
    =
    \begin{bmatrix}
        106.94^{\circ} \\
        -0.43^{\circ}  \\
        -0.38^{\circ}
    \end{bmatrix}.
\end{equation}

\subsection{Objects Detection Performance}
Table \ref{tab:model-map} and Table \ref{tab:model-ar} show accuracy results for the
retrained weights on the test set. After deploying to Jetson Nano, the model runs at an average processing time of 24 milliseconds, equivalent to 41.67 frames per second, and SSL objects can be detected for up to 3.5 meters distances using the onboard camera.

\begin{table}[ht]
    \setlength{\tabcolsep}{4pt}
    \centering
    \caption{Average Precision (AP) evaluation on the test set. Results are
        measured by Intersection over Union (IoU) threshold, where $50$ and $75$
        indexes indicate 0.5 and 0.75 IoU, while $\textbf{AP}_{S}$,
        $\textbf{AP}_{M}$ and $\textbf{AP}_{L}$ represent small, medium and
        large objects.}
    \label{tab:model-map}
    \begin{tabular}{ccccccc}
        \hline
        Model              & $\textbf{AP}$     & $\textbf{AP}_{50}$ &
        $\textbf{AP}_{75}$ & $\textbf{AP}_{S}$ & $\textbf{AP}_{M}$  &
        $\textbf{AP}_{L}$                                                        \\
        \hline
        SSD MobileNet v2   & $62.2\%$          & $93.4\%$           & $68.2\%$ &
        $35.0\%$           & $81.9\%$          & $91.1\%$                        \\
        \hline
    \end{tabular}
\end{table}
\vspace{-10pt}
\begin{table}[ht]
    \setlength{\tabcolsep}{4pt}
    \centering
    \caption{Average Recall (AR) evaluation on the test set. $1$ and $10$
        indexes indicate the maximum number of objects per image, while
        $\textbf{AR}_{S}$, $\textbf{AR}_{M}$ and $\textbf{AR}_{L}$ represent
        small, medium and large objects.}
    \label{tab:model-ar}
    \begin{tabular}{cccccc}
        \hline
        Model             & $\textbf{AR}_{1}$ & $\textbf{AR}_{10}$ &
        $\textbf{AR}_{S}$ & $\textbf{AR}_{M}$ & $\textbf{AR}_{L}$               \\ \hline
        SSD MobileNet v2  & $47.3\%$          & $68.8\%$           & $48.5\%$ &
        $85.6\%$          & $93.3\%$                                            \\
        \hline
    \end{tabular}
\end{table}

\subsection{Ball Localization}
\begin{figure}[ht]
    \begin{center}
        \includegraphics[width=0.8\textwidth]{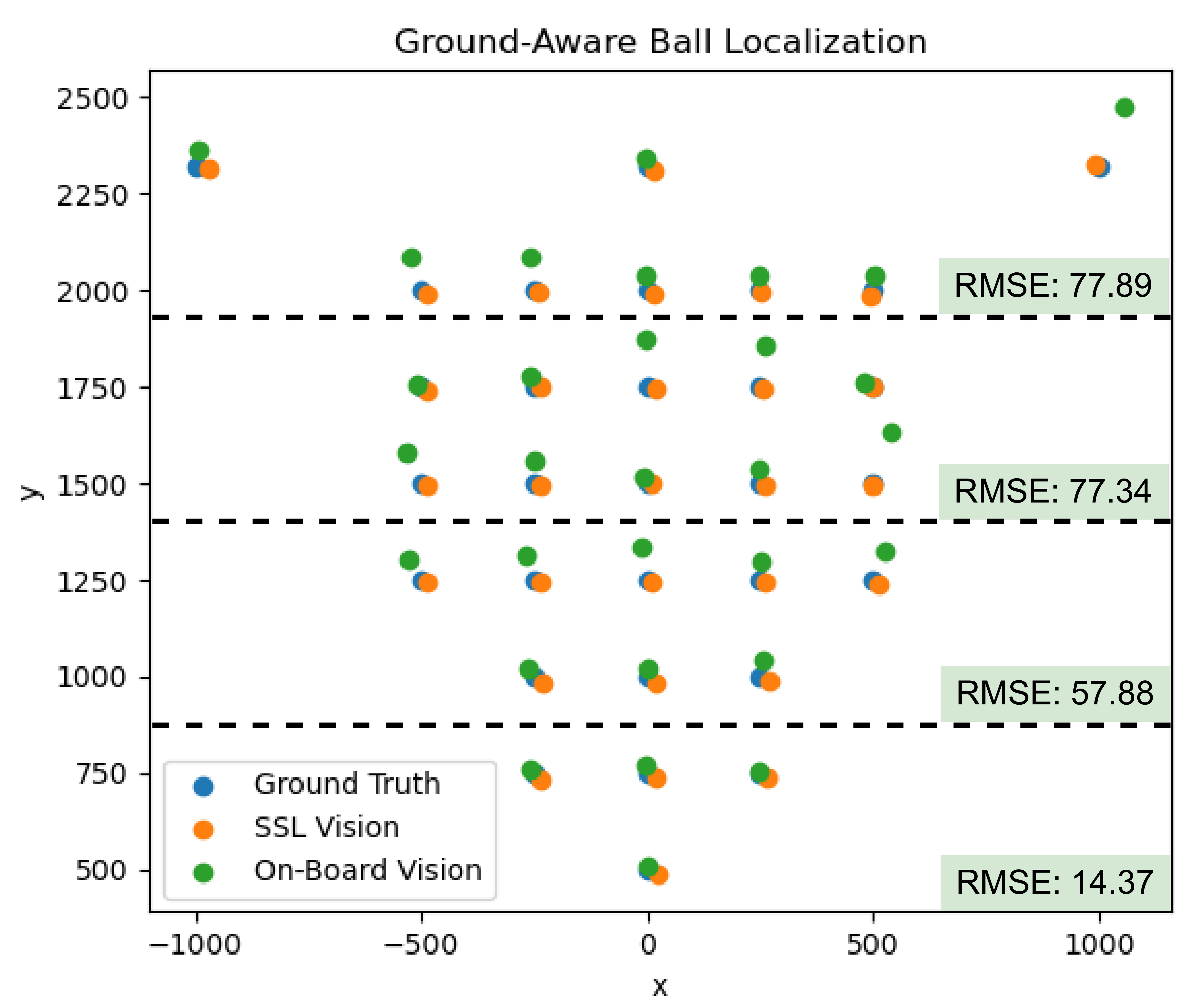}
    \end{center}
    \caption{Ball relative positions were estimated from the on-board detection system (green), being compared to SSL Vision (orange) and the ground truth (blue). Subsets of points were selected and the respective RMSE's were computed, reporting a 14.37mm error for the four nearest points. The dashed lines highlight the divisions between subsets and the RMSE for the whole set of points was 67.32 mm.}
    \label{fig:scatter-plot}
\end{figure}

For evaluating the proposed object localization approach, we chose 30 different
field positions for the ball. Since errors from points closer to the robot have
a higher impact than further ones for autonomous navigation, points were split
into subsets according to their distance. The estimated locations from the
onboard vision system and RMSE can be seen in Figure \ref{fig:scatter-plot}.

From the given plot and RMSE measurements, errors increase with the distance to
the robot. This is due to objects on screen getting smaller for further
distances, resulting in less accurate bounding boxes and position estimation
being more sensitive to pixel differences for distant points. Thus, we present
an in-depth analysis from the four nearest positions in Table
\ref{tab:evaluation}, showing that the proposed solution is capable of
overcoming SSL Vision accuracy for locations near the robot. Angles are measured by the tangent arc of relative XY positions.

\begin{table}[H]
    \setlength{\tabcolsep}{8pt}
    \centering
    \caption{Comparison between estimated and ground-truth relative positions and rotation. In the last two lines, mean error and standard deviation of our method are compared to SSL Vision for the correspondent set of points. Coordinates are measured in millimeters, while angles are in degrees.}
    \label{tab:evaluation}
    \begin{tabular}{ccc|ccc}
        \hline
        \multicolumn{3}{c|}{\textbf{Ground Truth} ($x$, $y$, $\theta$)} &
        \multicolumn{3}{c}{\textbf{On-Board Vision} ($x$, $y$, $\theta$)}          \\
        \hline
        0                                                               & 500    &
        0
                                                                        & -0.03
                                                                        & 508.86
                                                                        &
        0.00\textdegree
        \\
        \hline
        -250                                                            & 750    &
        -18.43
                                                                        &
        -259.17
                                                                        & 762.23
                                                                        &
        -18.78\textdegree
        \\
        \hline
        0                                                               & 750    &
        0
                                                                        & -1.08
                                                                        & 772.20
                                                                        &
        -0.08\textdegree
        \\
        \hline
        250                                                             & 750    &
        18.43
                                                                        & 247.12
                                                                        & 753.32
                                                                        &
        18.16\textdegree
        \\
        \hline
        \multicolumn{3}{c|}{Mean Error (Ours)}                          &
        \textbf{-0.03\textpm 3.39}                                      &
        \textbf{3.32\textpm 8.38}
                                                                        &
        \textbf{0.00\textdegree\textpm 0.15\textdegree}
        \\
        \hline
        \multicolumn{3}{c|}{Mean Error (SSL Vision)}                    &
        15.31\textpm 11.04                                              &
        -11.03\textpm 7.00                                              &
        0.72\textdegree\textpm 1.12\textdegree                                     \\ \hline
    \end{tabular}
\end{table}


Accuracy results can be interpreted from comparisons to the robot and ball dimensions: SSL robots have an approximate 100 mm wide front area for grabbing and shooting the ball, which has a fixed diameter of 42.7 mm. Mean errors of 3.32 mm, in the y axis, -0.03 mm, in the x axis, and 0\textdegree, in direction, should suffice the accuracy needed for autonomously approaching the ball, for instance.

\section{Conclusion} \label{ch5}

This work presents an approach for detecting and locating objects on the SSL soccer field, employing SSD MobileNet v2 CNN architecture for Object Detection and estimating ground points for calculating relative positions. Images from a chessboard pattern are used for computing intrinsic camera parameters. The camera pose is measured by solving a PnP problem for a set of 2D-3D correspondences from points on the field. The system is implemented on an NVIDIA Jetson Nano Developer Kit, enabling TensorRT acceleration. CNN-based object detection is employed, and we fit a linear regression model for predicting ground points based on bounding box positions. The pinhole camera model equation is solved for $z_{w} = 0$, and the object's relative XY coordinates are regressed.

Evaluation shows that our approach overcomes the current vision system accuracy for points near the camera with an average processing speed of 30 frames per second, while also respecting the league's 180 mm diameter restriction and consuming low power. Ball relative localization can be used for autonomously moving on its direction or rotating around it, for instance. In addition, our method is capable of detecting robots and goals, differently from existing ones in the SSL, while also being more robust to environment changes.

Future work include reproducing the presented procedure for estimating the robot and goal positions and searching for field lines detection solutions, which can have their relative coordinates calculated by the proposed method as well, enabling online camera pose calibration. Also, performance evaluations in a more dynamic environment, with moving objects and camera, must be done.


%
%
%
\bibliographystyle{splncs04}

\bibliography{references}

\end{document}